\documentclass[preprint,12pt]{elsarticle}

\usepackage{amssymb}

\journal{Pattern Recognition}

\begin{document}

\begin{frontmatter}



\title{A Review of Meta-Reinforcement Learning \\for Deep Neural Networks Architecture Search}

\author[label1,label2,label3]{Yesmina Jaafra}
\author[label2]{Jean Luc Laurent}
\author[label1]{Aline Deruyver}
\author[label3]{Mohamed Saber Naceur}
\address[label1]{ICube Laboratory, Université de Strasbourg, 300 bd Sébastien Brant, 67412 Illkirch, France}
\address[label2]{Segula Technologies, Parc d’activité de Pissaloup, 8 avenue Jean d’Alembert, 78190 Trappes, France}
\address[label3]{LTSIRS Laboratory, ENIT, 1002 Tunis, Tunisie }

\author{}

\address{}

\begin{abstract}
Deep Neural networks are efficient and flexible models that perform well for a variety of tasks such as image, speech recognition and natural language understanding. In particular, convolutional neural networks (CNN) generate a keen interest among researchers in computer vision and more specifically in classification tasks. CNN architecture and related hyperparameters are generally correlated to the nature of the processed task as the network extracts complex and relevant characteristics allowing the optimal convergence. Designing such architectures requires significant human expertise, substantial computation time and doesn’t always lead to the optimal network. Model configuration topic has been extensively studied in machine learning without leading to a standard automatic method. This survey focuses on reviewing and discussing the current progress in automating CNN architecture search.
\end{abstract}


\end{frontmatter}


\section{Introduction}
\label{sec:intro}

"A neuron is nothing more than a switch with information input and output. The switch will be activated if there are enough stimuli of other neurons hitting the information input. Then, at the information output, a pulse is sent to, for example, other neurons " \cite{Kriesel2007}. Brain-inspired machine learning imitates in a simplified manner the hierarchical operating mode of biological neurons \cite{SzeCYE17}.  The concept of artificial neural networks (ANN) achieved a huge progress from its first theoretical proposal in the 1950s until the recent considerable outcomes of deep learning. In computer vision and more specifically in classification tasks, CNN, which we will examine in this review, are among the most popular deep learning techniques since they are outperforming humans in some vision complex tasks \cite{Russakovsky2015}.

The origin of CNN that were initially established by  \cite{lecun1989} goes back to the 1950s with the advent of "perceptron", the first neural network prototyped by Frank Rosenblatt. However, neural network models were not extensively used until recently, after researchers overcame certain limits. Among these advances we can mention the generalization of perceptrons to many layers \cite{minsky69}, the emergence of backpropagation algorithm as an appropriate training method for such architectures \cite{rumelhart1986} and, mainly, the availability of large training datasets and computational resources to learn millions of parameters. CNN differ from classical neural networks in the fact that the connectivity of a hidden layer neuron is limited to a subset of neurons in the previous layer. This selective connection endow the network with the ability to operate, implicitly, hierarchical features extraction. For an image classification case, the first hidden layer can visualize edges, the second a specific shape and so on until the final layer that will identify the object.

CNN architecture consists of several types of layers including convolution, pooling, and fully connected. The network expert has to make multiple choices while designing a CNN such as the number and ordering of layers, the hyperparameters for each type of layer (receptive field size, stride, etc.). Thus, selecting the appropriate architecture and related hyperparameters requires a trial and error manual search process mainly directed by intuition and experience. Additionally, the number of available choices makes the selection space of CNN architectures extremely wide and impossible for an exhaustive manual exploration. Many research effort in meta-modeling tries to minimize human intervention in designing neural network architectures. In this paper, we first give a general overview and define the field of deep learning. We then briefly survey the history of CNN architectures. In the following section we review several methods for automating CNN design according to three dimensions:  search optimization, architecture design methods (plain or modular) and search acceleration techniques. Finally, we conclude the article with a discussion of future works.

\section{Background}
\label{sec:Background}
Before embarking with CNN, we will introduce in this section some basic generalities about artificial networks and deep learning.
\subsection{Artificial Neural Networks}
\begin{figure}[!ht]
\centering
  \includegraphics[width=0.6\linewidth]{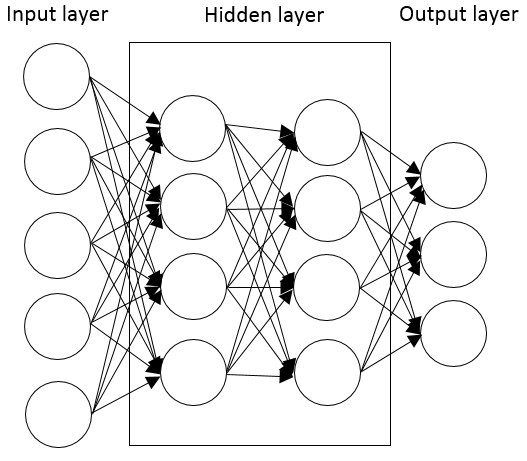}\\
  \caption{ Artificial neural network, containing an input layer, an output layer and two hidden layers.}\label{fig1}
\end{figure}

ANN are a major field of artificial intelligence that attempts to replicate human brain processing. Three types of neural layers distinguish an ANN: input, output and hidden layers. The latter operate transitional representations of the input data evolving from low level features (lines and edges) to higher ones (complex patterns) as far as deeper layers are reached. Figure \ref{fig1} provide an example of ANN involving classical fully-connected layers where every neuron is connected to all ones of the previous layer.

During training, an ANN aims at learning two types of parameters that will condition its predictive performance. First, connection weights that assess to which extent a neuron result will impact the output of higher level neuron. Second, the bias which is a global estimator of a feature presence across all inputs. Hence, a neuron output can be formalized through a linear combination of weighted inputs and associated bias:

\[output = (\sum_{i} input_i * weight_i)+bias\]

In order to allow the network operating non-linear transformations, an activation function is applied to the previous output. Equation \ref{equa1} presents an example of such transformation using one of the most common and efficient activation function which is the Rectified Linear Unit (ReLU) \cite{Lecun98}:

\begin{equation}
f(x) = max(x,0)
\label{equa1}	
\end{equation}

\subsection{Deep learning}
The concept of deep learning refers to machine learning processing within multi-layer ANN \cite{JarrettKRL09}. The training of these networks relies on a loss function evaluation. For example, in supervised learning the loss is assimilated to the matching accuracy between ANN predictions and real expected outputs. An iterative update procedure is implemented to adjust network parameters according to loss function computed gradient. This procedure is called Backpropagation since parameters updates are spread from final layers to initial ones. Deep learning implies a certain number of challenges such as vanishing/exploding gradient and overfitting. The solutions to these problems will be discussed when developing CNN design architectures in next sections.

\section{CNN Layers}
CNN are widely used in a great number of pattern and image recognition problems. Three main characteristics are making this deep learning technique successful and suitable to visual data. First, local receptive fields perfectly reflect image data specificity to be correlated locally and uncorrelated in global segments. Second, shared weights allows a substantial parameter reduction without altering image processing since the convolution is applicable to the whole image. Last, grid-structured image enable pooling operations that simplify data without losing useful information \cite{nielsen2018}.

\subsection{Convolutional Layer}
The convolutional layer is the basic CNN unit that has been inspired by physiological research evidence of hierarchical processing in the visual cortex of mammals \cite{hubel1962}. Simple cells detect primitive attributes while more compound structures are subsequently extracted by complex cells.
Thus, convolutional layer consists of a set of feature maps issued from convolving different filters (kernels) with an input image or previous layer output \cite{Krizhevsky2012}. The $2$-dimensional maps are stacked together to produce the resulting volume of the convolutional layer. This process reduces drastically the network complexity since the neurons of a same feature map share the same weights and bias maintaining a low number of parameters to learn \cite{Wu2015}.

The hyperparameters characterizing a convolutional layer are the depth $F$ (number of filters), the stride $S$ (filter movement from a receptive field to the next one) and the zero padding $P$ to control input size \cite{Goodfellow2016}. Assuming that the filter $size~(height,width,depth)=(h,w,D)$, the dimensions of the feature maps generated can be obtained according to:\\

$(H_1,W_1,F) = ((H+2P- h)/S+1,(W+2P-w)/S+1,F)$ \\
Where $(H,W,D)$ is the size (height, width, depth) of the input image. 

\subsection{Pooling Layer}

CNN architectures generally alternate convolution and pooling layers. The latter have the purpose of reducing network complexity and avoid the problem of overfitting. At biological level, pooling is assimilated to the behavior of cortical complex cells that reveal a certain degree of position invariance. A pooling layer neuron is connected to a region of the previous layer by performing a non-parameterized function. Thus it differs from convolution as it doesn't have learnable weights or bias and additionally, it keeps the same depth of the previous layer. Max pooling \cite{Zeiler2013a} is one of the most common type of pooling that consists in retaining the maximum value of a neurons cluster. It means that max pooling is detecting if a given feature has been identified in a receptive field without recording the exact location \cite{nielsen2018}.

\subsection{Fully connected Layer}
The convolution layers identify local features in the input data such as edges and shapes. The Fully connected layer operates the high level reasoning (classification for image case) by combining information from all the previous layers. As in a regular ANN, neurons at this level are fully connected to all ones in the previous layer. A softmax loss layer is then used to compute the probability distribution of the CNN final outputs.

\section{CNN Architecture History}
This section presents the most influential hand-crafted CNN architectures that have impacted the recent work on automatic architecture design. Most of them won at least one of the "ImageNet Large Scale Visual Recognition Competition" (ILSVRC) challenges \cite{Russakovsky2015}. 

\subsection{LeNet}
\begin{figure}[!ht]
\centering
  \includegraphics[width=0.9\linewidth]{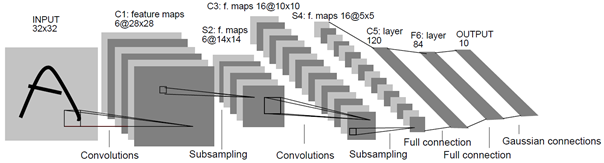}\\
  \caption{Architecture of LeNet-5 \cite{Lecun98}.}\label{fig2}
\end{figure}

As mentioned previously, LeNet \cite{Lecun98}  was the innovative work that introduced convolutional networks. The model was experimented successfully to classify handwritten digits without any preprocessing of the input image (of size $32*32$ pixels). LeNet architecture is illustrated in figure \ref{fig2}. It consists of an input and an output layers of respective sizes $32*32$ and $10$ as well as $6$ hidden Layers. The basic idea of this design is to operate multiple convolutions $(3)$ with pooling in-between $(2)$ then transmitting the final signal via a fully-connected layer toward the output layer. Unfortunately, due to the lack of adequate training data and computing power, it wasn't possible to extend this architecture to more complex applications.

\subsection{AlexNet}
\begin{figure}[!ht]
\centering
  \includegraphics[width=0.9\linewidth]{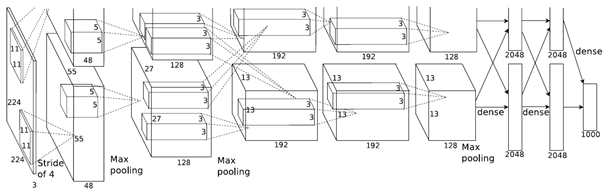}\\
  \caption{An illustration of AlexNet \cite{Krizhevsky2012}.}\label{fig3}
\end{figure}

AlexNet \cite{Krizhevsky2012} is one of the most influential deep CNN that won the ILSVRC (Imagenet Large Scale Visual Recognition Challenge) competitions in $2012$. As shown in figure \ref{fig3}, it is not much different from LeNet. Nevertheless, the corresponding architecture is deeper with $8$ layers in total, $5$ convolutional and $3$ fully connected. The effective contribution of AlexNet lies in several design and training specificities. First, it introduced the Rectified Linear Unit (ReLU) nonlinearity which helped to overcome the problem of vanishing gradient and boosted a faster training. Furthermore, AlexNet implements a dropout step \cite {Srivastava2014} that consists in setting to zero a predefined percentage of layers' parameters. This technique decreases learned parameters and controls neurons correlation in order to limit overfitting impact. Third, training process convergence is accelerated with momentum and conditional learning rate decrease (e.g. when learning stagnates). Finally, training data volume is increased artificially by generating variations of the original images that are shifted randomly. Thus the network learning is enhanced with the use of invariant representations of the data.

\subsection{VGGNet}
Submitted for the ILSVRC 2014, VGGNet \cite{Simonyan14} won the second place and demonstrated that deeper architectures achieve better results. Indeed, with its $19$ hidden layers, it was much deeper than previous convolutional networks. In order to allow an increase in depth without an exponential growth of the parameters number, smaller convolution filters $(3*3)$ were used in all layers (e.g. lower size than the $11*11$ filters adopted in AlexNet). An additional advantage of using smaller filters consists in reducing overlapping scanned pixels which results in feature maps with more local details \cite{Zeilerb13}.

\subsection{GoogLeNet}
\begin{figure}[!ht]
\centering
  \includegraphics[width=0.7\linewidth]{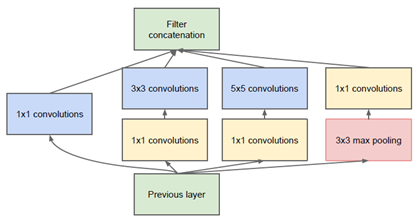}\\
  \caption{Inception module \cite{Szegedy2015}.}\label{fig5}
\end{figure}

Since it has been demonstrated that a CNN architecture size is positively correlated to its performance, recent efforts focus on how to increase the depth of a CNN while keeping an acceptable number of parameters. Winner of ILSVRC $2014$, GoogLeNet \cite{Szegedy2015} innovated network design by replacing the classical strategy of alternating convolutional and pooling layers with stacked Inception Modules depicted in figure \ref{fig5}. Despite being deeper than VGGNet with $22$ hidden layers, GoogLeNet requires outstandingly fewer parameters due to this sparse connection technique. Within an inception module, several convolutions with different scales and pooling are performed in parallel then concatenated in one single layer. This enables the CNN to detect patterns of various sizes within the same layer and avoid heavy parameters redundancies \cite{Szegedy2015}. GoogLeNet hidden layers consist of $3$ convolutions, $9$ inception blocks ($2$ layers deep each one) and one fully connected.

\subsection{ResNet}
\begin{figure}[!ht]
\centering
  \includegraphics[width=0.5\linewidth]{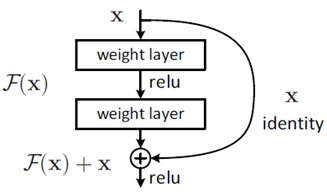}\\
  \caption{Residual learning: a building block \cite{He15}}\label{fig6}
\end{figure}

Deep Residual Network (ResNet) \cite{He15}, was the first neural network to exceed human-level accuracy in ImageNet Challenge (ILSVRC 2015). Thanks to residual connections, such kind of architecture went deeper and was implemented with multiple versions of $34$, $50$, $101$ and $152$ layers. Indeed, one of the difficulties with very deep networks training is the vanishing gradient during error backpropagation which penalizes the appropriate update of earlier layers weights. ResNet main contribution consists in dividing convolutional layers into residual blocks. Each block is bypassed by a residual (skip) connection that forwards the block input using an identity mapping. The final output is the summation of the block output and the mapped input as illustrated in figure \ref{fig6}.

By adding skip connections, backpropagation can be operated without any interference with previous layers which allows to prevent vanishing gradient and train very deep architectures. ResNet-$101$ consists of one convolutional layer followed by $33$ residual blocks ($3$ layers deep each one), and one fully connected layer.

\subsection{More Networks}
After ResNet \cite{He15} success, which exceeded human-level accuracy in (ILSVRC $2015$), the so-called modern hand-crafted CNN are still being designed on the basis of previous models looking for more efficiency and lower training time. Inception-v4 \cite{Szegedy16} is a new release of GoogLeNet that involved many more layers than the initial version. Inception-ResNet \cite{Szegedy16} is built as a combination of an Inception network and a ResNet, joining inception blocks and residual connections. The last example of this section is DenseNet (Dense Convolutional Networks) \cite{Huang16} where each dense block layer is connected via skip connections to all subsequent ones allowing the learning of new features.

\section{Meta-modeling for CNN automatic architecture design}
Meta-modeling for neural network architectures design aims at reducing the intervention of human expertise in this process. The earliest meta-modeling methods were based on genetic algorithms and Bayesian optimization then more recently, reinforcement learning became among the most implemented approaches \cite{Baker17}.

\subsection{Context of automation}
The performance of a neural network and particularly a CNN mainly depends on the setting of the model structure, the training process, and the data representation. All of these variables are controlled through a number of hyperparameters and impact the learning process to a large extent. In order to achieve an optimal performance of CNN, these hyperparameters including the depth of the network, learning rates, layer type, number of units per layer, dropout rates, etc., should be then carefully tuned. On the other hand, the advent of deeper and more complex modern architectures (see section $4$) is increasing the number and the types of hyperparameters. Hence, tuning step and more generally CNN architectures search become very expensive and heavy for an expert trial-and-error procedure.

Additionally, CNN parameters setting is considered as a black-box \cite{Bengio2013} optimization problem because of the unknown nature of the mapping between the architecture, the performance, and the learning task. In this context, automatic design solutions are highly required and instigates a large volume of research. The task of CNN hyperparameters tuning has been handled through meta-modeling that consists in applying machine learning models for designing CNN architectures. Three meta-modeling approaches are generally used in the literature of architecture search and will be described in the next paragraph: bayesian optimization, evolutionary algorithms and reinforcement learning.

\subsection{Meta-controllers}
Meta-modeling approaches perform iterative selection from the hyperparameters space and build associated architectures that are then trained and evaluated. Accuracies records are fed to meta-modeling controllers (meta-controllers) to guide next architectures sampling. Meta-controllers for CNN design are mainly based on bayesian optimization (\cite{Bergstra2011}, \cite{Domhan15}), evolutionary algorithms (\cite{Stanley2009}, \cite{Suganuma17}) or more recently on reinforcement learning (\cite{Zoph2017}, \cite{pham18a}).

Bayesian optimization is an efficient way to optimize black-box objective functions $f:X \to R$ that are slow to evaluate \cite{Brochu2010}. It aims at finding an input $x= arg \min_{x  \in X} f(x)$ that globally minimizes  $f$ where in the context of a machine learning algorithm, $x$ refers to the set of hyperparameters to optimize. The problem with this kind of optimization is that evaluating the objective function is very costly due to the great number of hyperparameters and the complex nature of models like deep neural networks. In order to overcome this problem, bayesian approaches propose probabilistic surrogate reconstruction of the objective function $p(f | D)$ where $D$ is a set of past observations. The evaluation of the empirical function is much cheaper than the true objective function \cite{Klein16}. Some of the most used probabilistic surrogate (regression) models are gaussian processes \cite{Rasmussen2005}, random forests \cite{Breiman2001} and tree-structured Parzen estimator \cite{Bergstra2011}.

Briefly, the processing of a bayesian optimization consists in building an empirical (probabilistic) model of the objective function. Then, iteratively, the model identifies a set of optimal hyperparameters for which the objective function returns corresponding results (e.g. loss values). Each feedback allows the update of the surrogate model and the guidance of hyperparameters predictions until the process reaches a termination condition.

Evolutionary algorithms present another strategy of hyperparameters optimization that modifies a set of candidate solutions (population) on the basis of a number of rules (operators). Following an iterative procedure of mutation, crossover and selection \cite{Eiben2015}, an evolutionary algorithm initializes, in a first step, a set of $N$ random networks to create a primary population. The second step consists in introducing a fitness function to score each network through its classification accuracy and keep the top ranked networks to construct the next generation. The evolutionary process continues until a termination criteria is met, which is generally defined as the maximum number of allowed generations. One of the advantage of evolutionary algorithms is the adaptation to complex combination of discrete (layer type) and continuous (learning rate) hyperparameters which is suitable to neuronal network optimization models \cite{Dufourq2017}.

\begin{figure}[!ht]
\centering
  \includegraphics[width=0.4\linewidth]{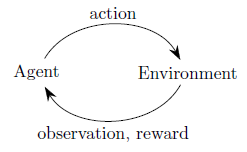}\\
  \caption{Illustration of the RL process.}\label{fig7}
\end{figure}

An important approach for goal-oriented optimization is reinforcement learning (RL) inspired from behaviorist psychology \cite{SuttonB98}. The frame of RL is an agent learning through interaction with its environment (figure \ref{fig7}). Thus the agent adapts its behavior (transition to a state $s_{t+1}$) on the basis of observed consequences (rewards) of an action $a_t$ taken in state $s_t$. The agent purpose is to learn a policy $\pi$ that is able to identify the optimal sequence of actions maximizing the expected cumulative rewards. The environment return reinforces the agent to select new actions to improve learning process, hence the name of reinforcement learning.

The methods developed to resolve reinforcement tasks are based on value functions, policy search or a combination of both strategies (actor-critic methods) \cite{Arulkumaran2017}. Value function methods consist in estimating the expected reward value $R$ when reaching a given state $s$ and following a policy $\pi$:\\

\begin{center}
$V^{\pi} (s) = \mathbb{E}[\Re|s,\pi$]
\end{center}

A recursive form of this function is particularly used in recent Q-learning \cite{Christopher1992} models assigned to CNN architecture design (\cite{Baker16}, \cite{Zhong2017}):

\begin{center}
$Q(s_t, a_t ) = Q(s_t, a_t ) + \alpha [r_{t+1}+ \gamma  max_{a} Q(s_{t+1}, a ) - Q(s_t , a_t)]$
\end{center}

Where $s_t$ is a current state, $a_t$ is a current action, $\alpha$ is the learning rate, $r_{t+1}$ is the reward earned when transitioning from time $t$ to the next and $\gamma$ is the discount rate.

In contrast to value function methods, policy search methods do not implement a value function and apply, instead, a gradient-based procedure to identify directly an optimal policy $\pi^*$. In this context, deep reinforcement learning is achieved when deep neural networks are used to approximate one of the reinforcement learning components : value function, policy or reward function \cite{Fuxiado2017}.

Among the active fields of designing CNN architectures through deep reinforcement learning, recurrent neural networks (RNN) arise as a valuable model that handles a set of tasks such as hyperparameters prediction (\cite{Zoph2017}, \cite{pham18a}). In fact, a RNN operates sequentially involving hidden units to store processing history, which allows the reinforcement learning to profit from past observations. Long short term memory networks (LSTM), a variant of RNN, offers a more efficient way of evolving conditionally on the basis of previous elements.

\section{Neural Architecture Search}

Various strategies have been developed to operate CNN architectures design for the majority of which reinforcement learning has been selected as meta-controller. This section is assigned to review in detail most recent promising automatic search approaches differentiated according to search spaces specificities and complexity level.

\subsection{Plain Architecture Design}

Some architecture search approaches focus on designing plain CNN which consists exclusively of conventional layers, mainly convolution, pooling and fully-connected. The resulting research space is relatively simple and the approaches contribution lies almost entirely in the design strategy.

\subsubsection{MetaQNN}

\begin{figure}[!ht]
\centering
  \includegraphics[width=0.65\linewidth]{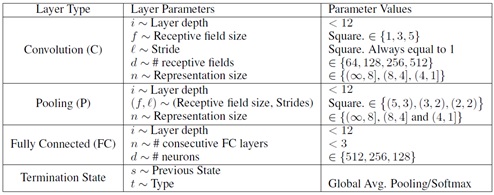}\\
  \caption{State space possible parameters \cite{Baker16}.}\label{fig8}
\end{figure}

MetaQNN model \cite{Baker16} relies on Q-learning, a type of reinforcement learning (refer to previous section for more details), to sequentially select network layers and their parameters among a finite space. This method implies, first, the definition of each learning agent state as a layer with all associated relevant parameters. As an example, $5$ layers are depicted in figure \ref{fig8}: convolution (C), pooling (P), fully connected (FC), global average pooling (GAP), and softmax (SM). 

\begin{figure}[!ht]
\centering
  \includegraphics[width=0.9\linewidth]{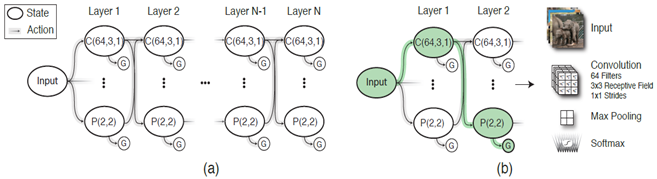}\\
  \caption{An illustration of the full state and action space (a) and a path that the agent has chosen (b) \cite{Baker16}.}\label{fig9}
\end{figure}

Second, the agent action space is assimilated to the possible layers the agent may move to given a certain number of constraints set intentionally, for the majority, to enable faster convergence. Figure \ref{fig9} illustrates a set of state and action spaces and an eventual agent path to design a CNN architecture. MetaQNN was evaluated competitive with similar and different hand-crafted CNN architectures as with existing automated network design methods.

\subsubsection{NAS}
\begin{figure}[!ht]
\centering
  \includegraphics[width=0.9\linewidth]{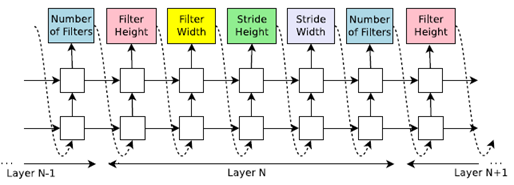}\\
  \caption{Illustration of the way the agent used to select hyperparameters \cite{Zoph2017}.}\label{fig10}
\end{figure}

Using reinforcement learning, \cite{Zoph2017} train a recurrent neural network to generate convolutional architectures. Figure \ref{fig10} shows a RNN controller generating sequentially CNN parameters associated to convolutional layers. Every sequence output is predicted by a softmax classifier then used as input of the next sequence. The parameters set consists of filter height and width, stride height and width and the number of filters per layer. The design of an architecture takes an end once the number of layers reaches a predefined value that increases all along training. The accuracy of the designed architecture is fed as a reward to train the RNN controller through reinforcement learning in order to maximize the expected validation accuracy of the next architectures. The experimentation of the global approach achieved competitive results on CIFAR-$10$ and Penn Treebank datasets.

\subsubsection{EAS}
\begin{figure}[!ht]
\centering
  \includegraphics[width=0.8\linewidth]{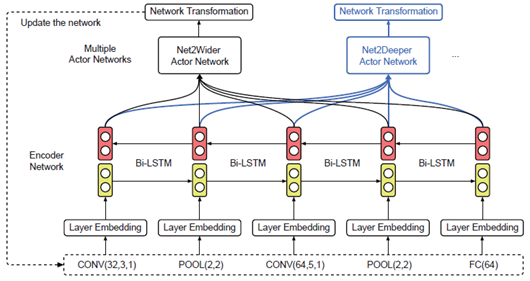}\\
  \caption{A meta-controller operation for network transformation \cite{Cai17a}.}\label{fig11}
\end{figure}

In their very recent work Efficient Architecture Search, \cite{Cai17a} implement network transformation techniques that allow reusing pre-existing models and efficiently exploring search space for automatic architecture design. This novel approach differs from the previous ones in the definition of reinforcement learning states and actions. The state is the current network architecture while the action involves network transformation operations such as adding, enlarging and deleting layers. Starting point architectures used in experiments are plain CNN which only consist of convolutional, fully-connected and pooling layers. EAS approach is inspired from Net2Net technique introduced in \cite{Chen15} and based on the idea of building deeper student network to reproduce the same processing of an associated teacher network. As shown in figure \ref{fig11}, an encoder network implemented with bidirectional recurrent neural network \cite{Schuster97} feeds actors network with given architectures. The selected actor networks performs $2$ types of transformation: widening layers in terms of units and filters and inserting new layers. EAS outperforms similar state-of-the-art models designed either manually or automatically with the attractive advantage of using relatively much smaller computational resources.

\subsection{Modular Architecture Design}

Most of recent work on neural architecture search is based on more complex modular (multi-branch) structures inspired by modern architectures presented in section $4$. Rather than operating the tedious search over entire networks, this second set of approaches focus on finding building blocks similarly to the ones used in, e.g. GoogLeNet and ResNet models. These multi-branch elements are then stacked repetitively involving skip connections to build the final deep architecture. As we will see through the models detailed in this section, "block-wise" architecture design reduces drastically search space speeding up search process, enhances generated networks performance and gives them more transferable ability through minor adaptation.

\subsubsection{BlockQNN}

\begin{figure}[!ht]
\centering
  \includegraphics[width=0.7\linewidth]{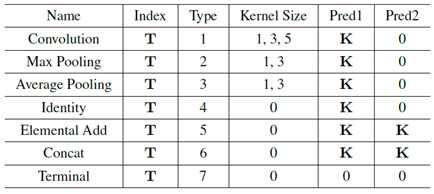}\\
  \caption{Network structure code space \cite{Zhong2017}.}\label{fig12}
\end{figure}

\begin{figure}[!ht]
\centering
  \includegraphics[width=0.7\linewidth]{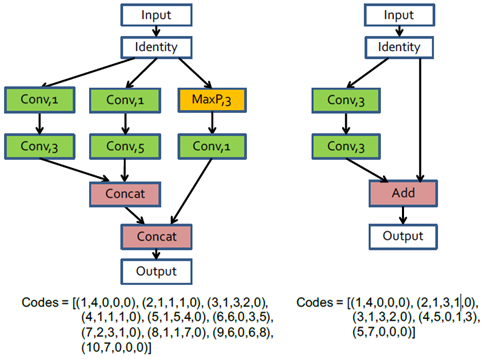}\\
  \caption{Representative block exemplars with their network structure codes \cite{Zhong2017}.}\label{fig13}
\end{figure}

One of the first approaches implementing block-wise architecture search, BlockQNN \cite{Zhong2017} automatically builds convolutional neworks using Q-Learning reinforcement technique \cite{Watkins89} with epsilon-greedy as exploration strategy \cite{Volodymyr2015}. The block structure is similar to ResNet and Inception (GoogLeNet) modern networks since it contains shortcut connections and multi-branch layer combinations. The search space of the approach is reduced given that the focus is switched to explore network blocks rather than designing the entire network. The block search space is detailed in figure \ref{fig12} and consists of $5$ parameters: a layer index (its position in the block), an operation type (selected among $7$ types commonly used), a kernel size and $2$ predecessors layers indexes. Figure \ref{fig13} depicts $2$ different samples of blocks, one with multi-branch structure and the second showing a skip connection. As described in previous sections, the Q-learning model includes an agent, states and actions, where the state represents the current layer of the agent and the action refers to the transition to the next layer. On the basis of defined blocks, the complete network is constructed by stacking them sequentially $N$ times.

\subsubsection{PNAS}

\begin{figure}[!ht]
\centering
  \includegraphics[width=0.8\linewidth]{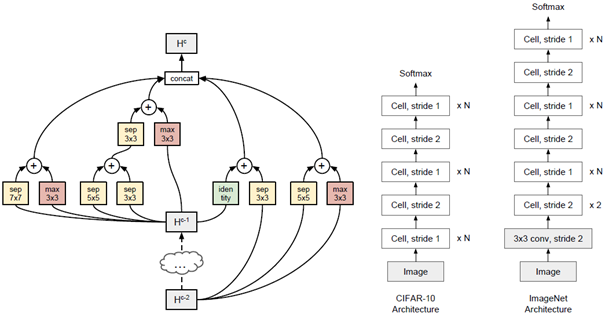}\\
  \caption{The best selected cell architecture of PNAS \cite{lui2017}. }\label{fig14}
\end{figure}

Progressive neural architecture search \cite{lui2017} proposes to explore the space of modular structures starting from simple models then evolving to more complex ones, discarding underperforming structures as learning progresses. The modular structure in this approach is called a cell and consists of a fixed number of blocks. Each block is a combination of $2$ operators among $8$ selected ones such as identity, pooling and convolution. A cell structure is learned first then it's stacked $N$ times in order to build the resulting CNN. The main contribution of PNAS lies in the optimization of the search process by avoiding direct search in the entire space of cells. This was made possible with the use of a sequential model-based optimization (SMBO) strategy. The initial step consists in building, training and evaluating all possible $1$-block cells. Then the cell is expanded to $2$-block size exploding the number of total combinations. The innovation brought by PNAS is to predict the performance of the second level cells by training a RNN (predictor) on the performance of previous level ones. Only the $K$ best cells (i.e. most promising ones) are transferred to the next step of cell size expansion. This process is repeated until the maximum allowed blocks number is reached. With an accuracy comparable to NAS \cite{Zoph2017} approach, PNAS is up to $5$ times faster using a cell maximum size of $5$ blocks and $K$ equal to $256$. This result is due to the fact that performance prediction takes much less time than full training of designed cells. The best cell architecture is shown in figure \ref{fig14}.

\subsubsection{ENAS}

\begin{figure}[!ht]
\centering
  \includegraphics[width=0.65\linewidth]{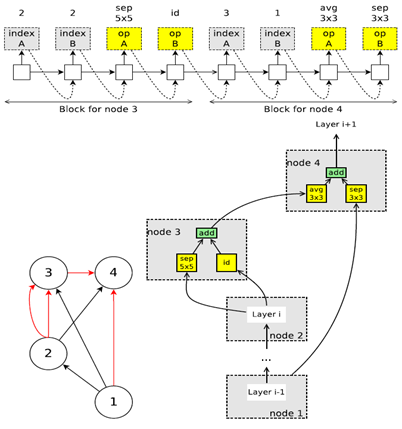}\\
  \caption{Illustration of 4-nodes cell \cite{pham18a}.}\label{fig15}
\end{figure}

Efficient neural architecture search \cite{pham18a} comes in the continuity of previous work NAS \cite{Zoph2017} and PNAS \cite{lui2017}. It explores a cell-based search space through a controller RNN trained with reinforcement learning. The cell structure is similar to PNAS model where block concept is replaced with a node that consists of $2$ operations and two skip connections. The controller RNN manages thus $2$ types of decisions at each node. First it identifies $2$ previous nodes to connect to, allowing the cell to set skip connections. Second, the controller selects $2$ operations to implement among a set of $1$ identity, $2$ depth wise-separable convolutions of filter sizes $3*3$ and $5*5$ \cite{Chollet16}, max pooling and average pooling both of size $3*3$. Within each node, the operations results are added in order to constitute an input for the next node. Figure \ref{fig15} illustrates the design of a $4$-node cell. At the end, the entire CNN is built by stacking $N$ times convolutional cells. 

Another contribution of ENAS consists in sampling mini-batches from validation dataset to train designed models. The models with the best accuracy are then trained on the entire validation dataset. Additionally, the approach efficiency is greatly improved by implementing a weight sharing strategy. Each node has its own parameters (used when involved operations are activated) that are shared through inheritance by the generated child models. The latter are hence not trained from scratch saving a considerable processing time. ENAS provides competitive results on CIFAR-$10$ and Penn Treebank datasets. It specifically takes much less time to build the convolution cells than previous approaches that adopt the same strategy of designing modular structures then stack them to obtain a final CNN.

\subsubsection{EAS With Path Level Transformation}

\begin{figure}[!ht]
\centering
  \includegraphics[width=0.85\linewidth]{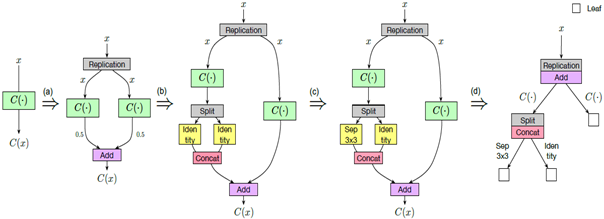}\\
  \caption{Path-level transformation: from a single layer to a tree-structured motif \cite{Cai2018b}.}\label{fig16}
\end{figure}

A developed version of EAS \cite{Cai17a} which adopts network transformation for efficient CNN architecture search is presented in \cite{Cai2018b}. The new approach tackle the constraint of only performing plain architecture modification (layer-level), e.g. adding (removing) units, filters and layers, by using path-level transformation operations. The proposed model is similar to (\cite{Cai17a}) where the reinforcement learning meta-controller samples network transformation actions to build new architectures. The latter are then trained and resulting accuracies are used as reward to update the meta-controller. However, certain changes have been implemented in order to adapt search methods to the tree-structured architecture space: using a tree-structured LSTM, (\cite{Tai15}) as meta-controller, defining a new action space consisting of feature maps allocation schemes (replication, skip), merge schemes (add, concatenation, none) and primitive operations (convolution, identity, depthwise-separable convolution, etc.). Figure \ref{fig16} presents an example of transformation decisions operated by the meta-controller. Experimenting with ResNet and DenseNet architectures as base input, the path level transformation approach achieves competitive performance with state-of-the-art models maintaining low computational resources comparable to EAS approach ones.

\subsection{Architecture search accelerators}

Reinforcement learning methods have been applied successfully to design neural networks. Although multi-branch structures and skip connections improves the efficiency of architectures automatic search, the latter is still computationally expensive (hundreds of GPU hours), time consuming and requires further acceleration of learning process. Thus, in addition to the methods assigned to architectural search optimization and complex component building, some techniques are developed to speed up learning and are depicted in the current section.

Early stopping strategy proposed in \cite{Zhong2017} enables fast convergence of the learning agent while maintaining an acceptable level of efficiency. This is possible by taking into account intermediate rewards ignored in previous works (set to zero delaying reinforcement learning convergence \cite{SuttonB98}. In such case, the agent stops searching in an early training phase as the accuracy rewards reach higher levels in fewer iterations. The reward function is redefined in order to include designed block complexity and density and avoid possible poor accuracy resulting from training early stopping.
 
A second technique is presented in \cite{Zhong2017} which consists of a distributed asynchronous framework assembling $3$ nodes with different functions. The master node is the place where block structures are sampled by agent. Then, in the controller node, the entire network is built from generated blocks and transmitted to multiple compute nodes for training. The framework is a kind of simplified parameter-server \cite{Dean2012} and allows the parallel training of designed networks in each compute nodes.  Hence, the whole design and learning processing is operated in multiple machines and GPUs. \cite{Zoph2017} uses the same parameter server scheme with replication of controllers in order to train various architectures in parallel.

As seen previously, reinforcement learning policies use explored architectures performance as a guiding reward for controllers updates. Training and evaluating every sampled architecture (among hundreds) on validation data is responsible for most of computational load. Extracting architecture performance was consequently subject to several estimation attempts. A number of approaches focus on performance prediction on the basis of past observations. Most of such techniques are based on learning curve extrapolation \cite{Domhan15} and surrogate models using RNN predictor \cite{Lui2017b} that aim at predicting and eliminating poor architectures before full training. Another idea to estimate performance and rank designed architectures is to use simplified (proxy) metrics for training such as data subsets (mini-batches)  \cite{pham18a} and down-sampled data (like images with lower resolution) \cite{Hinz2018Speeding}. 

Network transformation is one of the more recent techniques assigned to accelerate neural architecture search (\cite{Cai2018b}, \cite{elsken2018a}). It consists in training explored architectures reusing previously trained or existing networks. This modeling feature allows to address a limitation of reinforcement learning approaches where training is performed with a random initialization of weights. Thus, extending network morphisms \cite{Wei17} to initiate architecture search through the transfer of experience and knowledge reflected by reused weights enables the framework to scrutinize the search space efficiently.

Although the techniques presented above have saved substantial computational resources for neural architecture search, there is still more effort needed to examine the extent of bias impact of such techniques on the search process. Indeed, it's crucial to assure that modifications brought through re-sampled data, discarded cases and early convergence do not influence the models original predictions. Further studies are thus required to verify that learning accelerators do not have amplified effect on approaches predictions and validation accuracies.

\section{Conclusion}
The review of recent work trend on automatic design of CNN architectures raised some methodological options that are adopted by the majority of built approaches. Despite some attempts to use design meta-controllers based on evolutionary algorithms (\cite{Stanley2009}, \cite{Suganuma17}) and Bayesian optimization (\cite{Domhan15}, \cite{Mendoza2016}), reinforcement learning has shown promising empirical results and stands as the preferred strategy to train design controllers \cite{Perez2018}. 

Another common conception option is the introduction of multi-branch (modular) structures as an elementary component of the entire network which restricts the search space to block/cell level. The plain network design is generally kept as a first step of proposed approaches application (\cite{Zoph2017},  \cite{Cai17a}) given that it leads to simple networks and allows to focus on the method itself before switching to more complex structures with modular design (\cite{pham18a}, \cite{Cai2018b}). A third option used in design approaches at a lower scale is the prediction of explored architectures rewards before full training the most promising ones (\cite{Domhan15}, \cite{pham18a}). This training acceleration technique is implemented for performance improvement purpose and requires further attention to control possible bias impact on the models behavior.

The success of current reinforcement-learning-based approaches to design CNN architectures is widely proven especially for image classification tasks.  However, it is achieved at the cost of high computational resources despite the acceleration attempts of most of recent models. Such fact is preventing individual researchers and small research entities (companies and laboratories) from fully access to this innovative technology \cite{Cai17a}. Hence, deeper and more revolutionary optimizing methods are required to practically operate CNN automatic design.  Transformation approaches based on extended network morphisms \cite{Cai2018b} are among the first attempts in this direction that achieved drastic decrease in computational cost and demonstrated generalization capacity. Additional future directions to control automatic design complexity is to develop methods for multi-task problems \cite{Liang2018} and weights sharing \cite{bender18} in order to benefit from knowledge transfer contributions.



\bibliographystyle{elsarticle-num} 





\end{document}